\documentclass[sigconf,cameraready]{acmart}

\usepackage{booktabs}
\usepackage{graphicx}
\usepackage{multirow}
\usepackage{xcolor}
\usepackage{subcaption}
\usepackage{amsmath}
\usepackage{tikz}
\usetikzlibrary{positioning, arrows.meta, fit, backgrounds, calc}

\setcopyright{none}
\renewcommand\footnotetextcopyrightpermission[1]{}

\begin{document}

\title{CAGE: Bridging the Accuracy-Aesthetics Gap in Educational Diagrams via Code-Anchored Generative Enhancement}

\author{Dikshant Kukreja}
\authornote{Equal contribution}
\affiliation{
  \institution{IIIT Delhi, India}}
\email{dikshant22176@iiitd.ac.in}

\author{Kshitij Sah}
\authornotemark[1]
\affiliation{
  \institution{IIIT Delhi, India}}
\email{kshitij22256@iiitd.ac.in}

\author{Karan Goyal}
\affiliation{%
  \institution{IIIT Delhi, India}}
\email{karang@iiitd.ac.in}

\author{Mukesh Mohania}
\affiliation{%
  \institution{IIIT Delhi, India}}
\email{mukesh@iiitd.ac.in}

\author{Vikram Goyal}
\affiliation{%
  \institution{IIIT Delhi, India}}
\email{vikram@iiitd.ac.in}

\renewcommand{\shortauthors}{Kukreja and Sah et al.}

\begin{abstract}
Educational diagrams---labeled illustrations of biological processes, chemical structures, physical systems, and mathematical concepts---are essential cognitive tools in K--12 instruction. Yet no existing method can generate them both accurately and engagingly. Open-source diffusion models produce visually rich images but catastrophically garble text labels. Code-based generation via LLMs guarantees label correctness but yields visually flat outputs. Closed-source APIs partially bridge this gap but remain unreliable and prohibitively expensive at educational scale. We quantify this \emph{accuracy--aesthetics dilemma} across all three paradigms on 400 K--12 diagram prompts, measuring both label fidelity and visual quality through complementary automated and human evaluation protocols. To resolve it, we propose \emph{CAGE} (\textbf{C}ode-\textbf{A}nchored \textbf{G}enerative \textbf{E}nhancement): an LLM synthesizes executable code producing a structurally correct diagram, then a diffusion model conditioned on the programmatic output via ControlNet refines it into a visually polished graphic while preserving label fidelity. We also introduce \emph{EduDiagram-2K}, a collection of 2,000 paired programmatic--stylized diagrams enabling this pipeline, and present proof-of-concept results and a research agenda for the multimedia community.
\end{abstract}

\ccsdesc[500]{Computing methodologies~Computer vision}
\ccsdesc[300]{Computing methodologies~Machine learning}

\keywords{CAGE, Educational diagram generation, Diffusion models, Code synthesis, Multimodal generation, K--12 education}

\maketitle

\section{Introduction}
\label{sec:intro}

In K--12 classrooms, diagrams are not decorative embellishments---they are \emph{cognitive scaffolds} that shape how students construct mental models of scientific and mathematical concepts~\cite{mayer2013multimedia, mayer2021evidence}. A labeled diagram of the human heart, a flowchart of photosynthesis, or a coordinate geometry illustration serves a fundamentally different purpose from a stock photograph: every label, every arrow, and every spatial relationship encodes factual information. When a label is wrong---when ``aorta'' is rendered as ``arota'' or ``mitochondria'' becomes an illegible smear---the diagram does not merely look bad. It teaches a child something false.

The automated generation of such diagrams has become an increasingly urgent need as artificial intelligence tools proliferate in educational settings. The global market for AI in K--12 education has grown at a compound annual growth rate of 38\%, yet the overwhelming majority of generative tools are designed for corporate or academic contexts~\cite{mcallister2026understanding}. Systems such as PPTAgent~\cite{zheng2025pptagent}, SlideGen~\cite{liang2025slidegen}, DocPres~\cite{bandyopadhyay2024enhancing}, and D2S~\cite{sun2021d2s} represent the state of the art in automated presentation generation, but they operate under a shared assumption: that images \emph{already exist} in a source document and need only be retrieved and positioned. None of these systems generate educational diagrams from scratch, and none address the stringent accuracy requirements of K--12 content. Meanwhile, the generative AI landscape offers two dominant paradigms for visual content creation, neither of which solves the educational diagram problem:

\begin{enumerate}
    \item \textbf{Diffusion-based image generation.} Models such as Stable Diffusion XL~\cite{podell2023sdxl}, Flux~\cite{greenberg2025demystifying}, and DALL\textperiodcentered E~3~\cite{betker2023improving} can produce visually compelling images from text prompts. However, a well-documented limitation of diffusion models is their consistent failure to render legible, accurate text within images~\cite{ma2023glyphdraw, zhang2024brush}. For educational diagrams, where textual labels carry the core pedagogical content, this limitation is disqualifying.

    \item \textbf{Code-based programmatic rendering.} Large language models can synthesize executable Python, \LaTeX{}/TikZ, or Mermaid code that renders diagrams with guaranteed label accuracy through standard graphics libraries~\cite{tian2024chartgpt, narechania2020nl4dv, luo2021natural}. However, the resulting outputs---Matplotlib plots, bare TikZ schematics---lack the visual polish and engagement quality that effective K--12 materials demand.
\end{enumerate}

Closed-source commercial APIs (DALL\textperiodcentered E~3, Gemini, GPT-4o native image generation) partially bridge this gap, achieving better---though still imperfect---text fidelity alongside higher visual quality. However, their per-image pricing renders them economically infeasible for resource-constrained schools, and their proprietary nature precludes reproducible research. In this paper, we make three contributions:

\begin{enumerate}
    \item \textbf{Empirical quantification of the accuracy--aesthetics dilemma.} We evaluate open-source diffusion models, LLM-driven code synthesis, and closed-source APIs on a benchmark of 400 educational diagram prompts across biology, chemistry, physics, and mathematics, measuring both label accuracy (via OCR-based exact-match and character error rate) and visual quality (via Fr\'{e}chet Inception Distance against a curated textbook reference set and human visual appeal ratings). Our results confirm that no existing single-stage approach delivers both accurate labels and visually engaging outputs (\S\ref{sec:dilemma}).

    \item \textbf{CAGE: a two-stage paradigm.} We propose decoupling accuracy from aesthetics by first synthesizing executable code that renders a structurally correct diagram, then applying a diffusion model conditioned on the programmatic output via structural controls (ControlNet with Canny edge conditioning) to produce a visually polished result while preserving label fidelity (\S\ref{sec:method}).

    \item \textbf{EduDiagram-2K: a paired dataset.} We introduce a novel dataset of approximately 2,000 paired programmatic--stylized educational diagrams spanning K--12 STEM subjects, enabling the training and evaluation of structure-preserving diagram refinement models (\S\ref{sec:dataset}).
\end{enumerate}

We conclude with a research agenda identifying open sub-problems that this paradigm surfaces for the multimedia community (\S\ref{sec:agenda}).

\section{The Accuracy--Aesthetics Dilemma}
\label{sec:dilemma}

\begin{figure*}[t]
    \centering
    \begin{subfigure}[t]{0.49\textwidth}
        \centering
        \includegraphics[width=\textwidth,height=6cm]{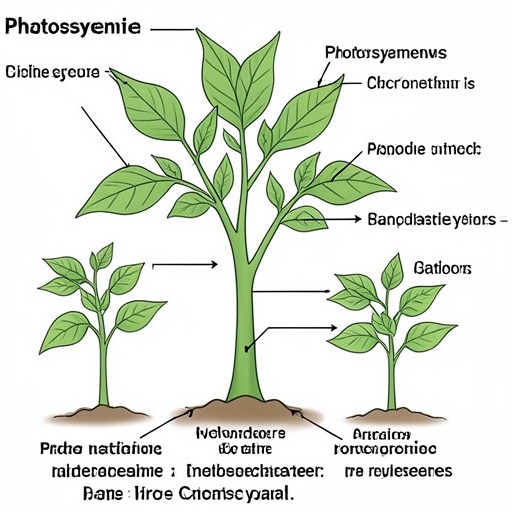}
        \caption{SDXL (open-source)}
    \end{subfigure}\hfill
    \begin{subfigure}[t]{0.49\textwidth}
        \centering
        \includegraphics[width=\textwidth,height=6cm]{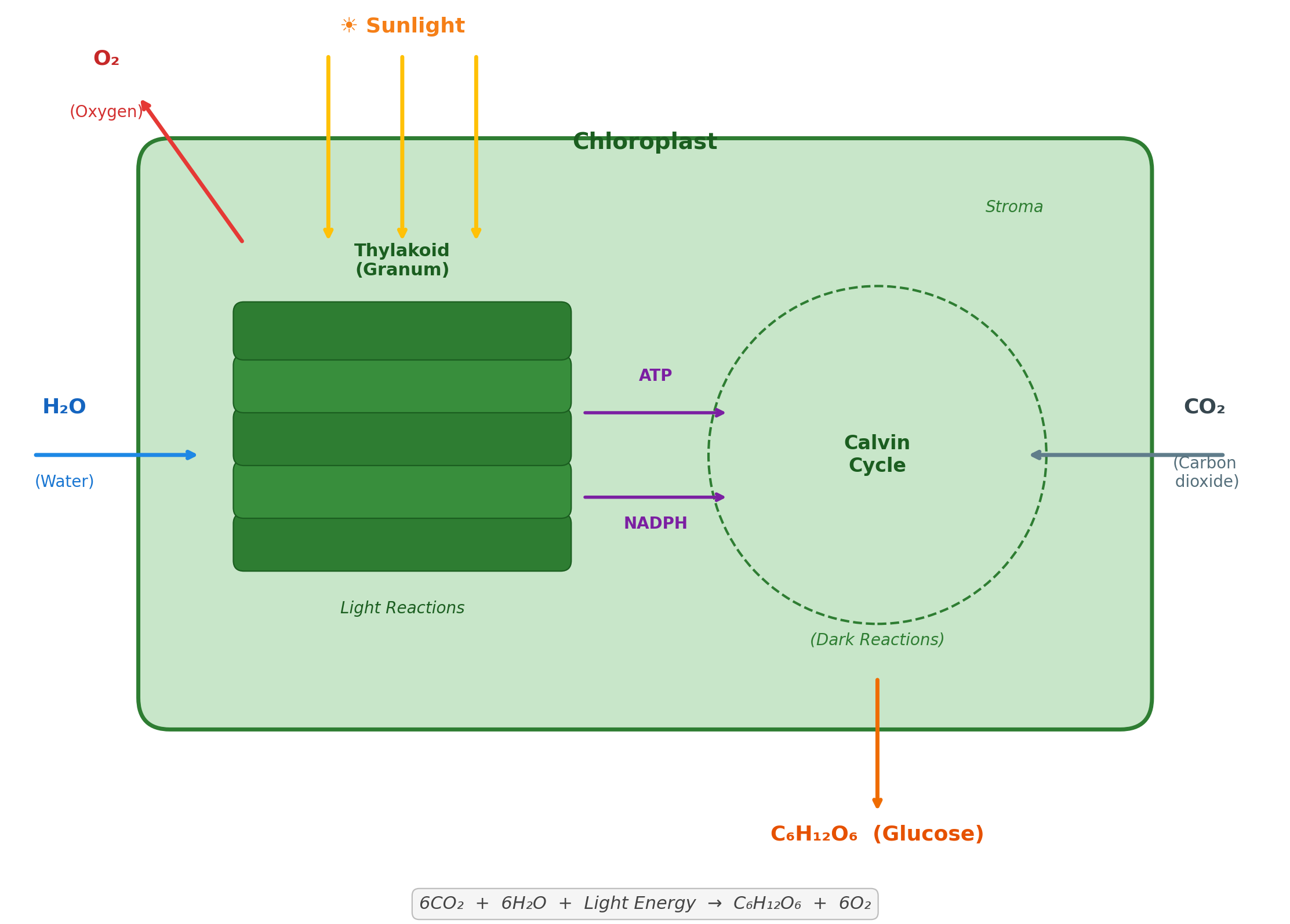}
        \caption{Code-based (LLM)}
    \end{subfigure}\hfill
    \begin{subfigure}[t]{0.49\textwidth}
        \centering
        \includegraphics[width=\textwidth]{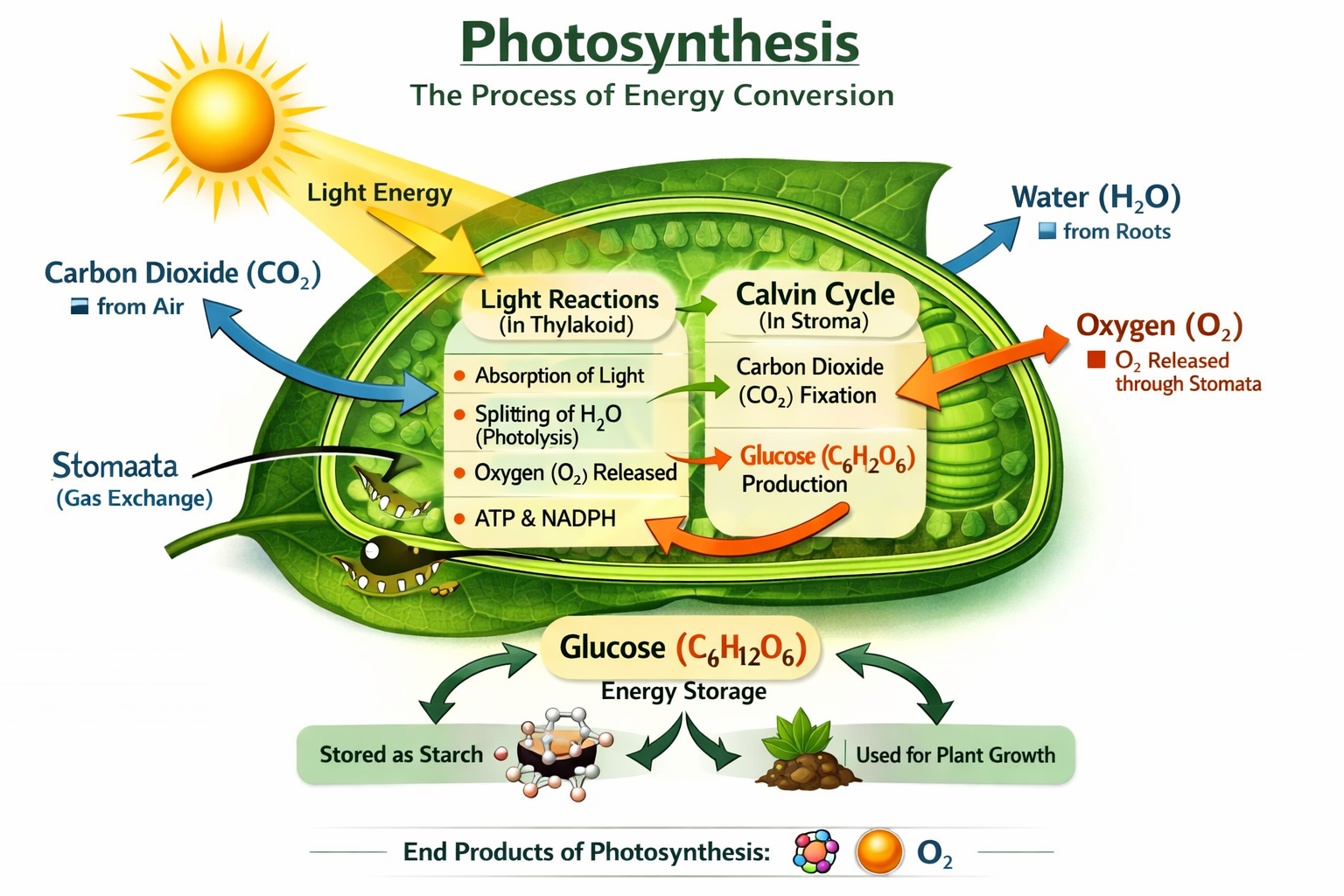}
        \caption{Gemini (closed-source)}
    \end{subfigure}\hfill
    \begin{subfigure}[t]{0.49\textwidth}
        \centering
        \includegraphics[width=\textwidth]{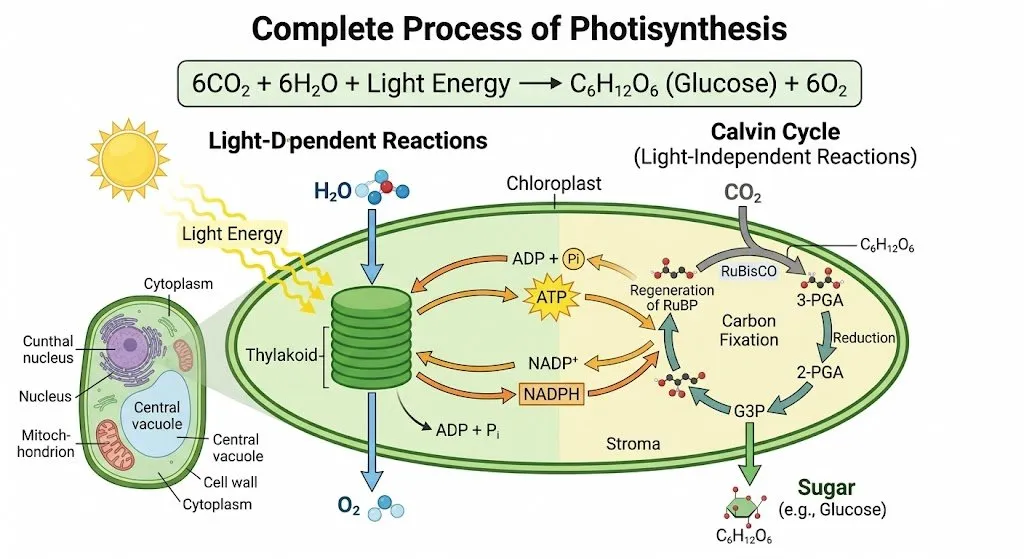}
        \caption{CAGE (ours)}
    \end{subfigure}
    \caption{The accuracy--aesthetics dilemma visualized on the prompt ``labeled diagram of photosynthesis.'' (a) Open-source diffusion: produces engaging visuals with catastrophically garbled labels. (b) Code-based generation: guarantees label accuracy but yields visually flat output. (c) Closed-source APIs: improve fidelity at significant per-image cost but remain imperfect (e.g., ``Stomaata''). (d) CAGE: achieves both accuracy and visual quality at zero marginal cost.}
    \label{fig:comparison}
\end{figure*}

The core claim of this paper---that educational diagram generation faces a fundamental tension unresolvable by any single existing approach---requires empirical grounding. In this section, we present a systematic evaluation of three paradigms on a common benchmark of 400 K--12 educational diagram prompts, measuring both label accuracy and visual quality. Please see Figure \ref{fig:comparison} for better understanding.

\subsection{Benchmark Design}
\label{sec:benchmark}

We construct a benchmark of 400 educational diagram prompts stratified across four STEM subjects: biology (110 prompts), chemistry (95), physics (95), and mathematics (100). Each prompt specifies the diagram topic and an explicit set of \emph{ground-truth labels} that must appear in the output. For example:

\begin{quote}
\small
\textit{``Generate a labeled diagram of the human digestive system showing: mouth, esophagus, stomach, liver, gallbladder, pancreas, small intestine, large intestine, rectum.''}
\end{quote}

This design enables automated evaluation: we apply OCR to each generated image and compute the fraction of ground-truth labels that are correctly rendered. Prompts are drawn from standard K--12 curriculum topics aligned with common core and NGSS standards, spanning elementary through high school grade bands.

We define two primary metrics for \textbf{label accuracy}:
\begin{itemize}
    \item \textbf{Label Exact-Match Rate (LEM):} The fraction of ground-truth labels that appear verbatim (case-insensitive) in the OCR-extracted text from the generated image.
    \item \textbf{Character Error Rate (CER):} The character-level edit distance between expected labels and their closest OCR-detected matches, normalized by the total expected character count.
\end{itemize}

To quantify \textbf{visual quality}, we employ two complementary aesthetic measures:
\begin{itemize}
    \item \textbf{Fr\'{e}chet Inception Distance (FID):} We compute FID~\cite{heusel2017gans} between each paradigm's generated outputs and a curated reference set of 500 textbook-quality educational diagrams sourced from openly licensed K--12 materials. Unlike standard FID evaluations that measure proximity to natural image distributions, our reference set captures the specific visual characteristics of professional educational illustrations---clean lines, purposeful color palettes, clear visual hierarchy, and age-appropriate complexity. Lower FID indicates closer distributional alignment with the textbook standard.
    \item \textbf{Human Visual Appeal (HVA):} A panel of 6 annotators with educational publishing or instructional design experience rates a stratified random subset of 200 generated diagrams (50 per paradigm) on a 5-point Likert scale. Each diagram is evaluated across four dimensions: (1)~color quality and contrast, (2)~professional appearance, (3)~visual engagement (``would a student want to look at this?''), and (4)~clarity of visual hierarchy. We report the mean composite score and inter-annotator agreement via Krippendorff's~$\alpha$.
\end{itemize}

\subsection{Open-Source Diffusion Models}
\label{sec:open_diffusion}

We evaluate three representative open-weight diffusion models: Stable Diffusion XL 1.0 (SDXL)~\cite{podell2023sdxl}, Flux.1-dev~\cite{greenberg2025demystifying}, and Stable Diffusion 3 Medium (SD3)~\cite{esser2024scaling}. Each model receives the same 400 prompts with default generation parameters. Table~\ref{tab:accuracy} (rows 1--3) reports the results.

The findings reveal a striking asymmetry. On the accuracy axis, all three models perform poorly: the mean label exact-match rate remains below 19\%, with chemistry prompts---which require subscripts, chemical formulas, and special characters---performing worst at 8.2\% LEM. However, on the aesthetics axis, these same models excel: they achieve the lowest (best) FID scores (95) and the highest HVA ratings (4.1/5), producing visually rich outputs with appealing colors, depth, and texture. This asymmetry is the accuracy--aesthetics dilemma in its sharpest form. Common accuracy failure modes include:

\begin{itemize}
    \item \textbf{Character-level garbling:} ``Mitochondria'' rendered as ``Mitochndira'' or ``Mitohcondra.''
    \item \textbf{Label fabrication:} Labels that do not correspond to any requested term appear in the diagram.
    \item \textbf{Complete omission:} Requested labels are absent, with the model generating unlabeled or incorrectly labeled regions.
    \item \textbf{Spatial incoherence:} Labels appear disconnected from the structures they should annotate.
\end{itemize}

These failures are not incidental. They reflect a well-characterized architectural limitation: diffusion models operating in a VAE-compressed latent space~\cite{rombach2022high} struggle with the high-frequency, precise spatial patterns required for legible text~\cite{ma2023glyphdraw, zhang2024brush}. While recent work on glyph-aware encoders~\cite{ma2023glyphdraw}, text-masked compositing~\cite{zhang2024brush}, and dedicated text-painting models~\cite{chen2023textdiffuser} has improved \emph{scene text} rendering in natural images, educational diagrams present a uniquely challenging setting in which text labels are the \emph{primary semantic content}, not a peripheral overlay.

\subsection{Code-Based Generation via LLMs}
\label{sec:code_gen}

We evaluate the same 400 prompts using LLM-driven code synthesis. Each prompt is reformulated as an instruction to generate executable Python code (using Matplotlib, NetworkX, or custom drawing libraries) that renders the requested diagram. We evaluate outputs from GPT-4o and Claude 3.5 Sonnet, executing the generated code and rendering the resulting image.

As expected, the label exact-match rate is near-perfect: 96\% LEM across all subjects, since labels are rendered via standard font-rendering libraries rather than learned generation. The structural correctness---connectivity of flowchart nodes, correct spatial arrangement of anatomical components---is also high, as it is determined by the code logic rather than stochastic sampling.

However, the visual quality of these outputs is categorically inferior to diffusion-generated images. Code-generated diagrams achieve FID scores of 188 against our textbook reference set---substantially worse than diffusion outputs---and receive HVA ratings of only 2.1/5. Matplotlib diagrams render with default styling: uniform colors, sans-serif labels, no texture or depth, no visual hierarchy. For K--12 education, where visual engagement is a documented factor in learning outcomes~\cite{mayer2013multimedia}, these outputs fail the \emph{aesthetics} criterion. A Matplotlib rendering of photosynthesis doesn't look like a textbook illustration but like a homework assignment.

\subsection{Closed-Source Commercial APIs}
\label{sec:closed_source}

We evaluate DALL\textperiodcentered E~3 (via the OpenAI API), Gemini's image generation (via the Google API), and GPT-4o's native image generation on the same 400 prompts. Table~\ref{tab:accuracy} reports label accuracy, visual quality, and per-image cost.

Closed-source models occupy an intermediate position on \emph{both} axes. They achieve substantially higher label accuracy than their open-source counterparts---DALL\textperiodcentered E~3 reaches 64\% LEM---but remain far from the near-perfect accuracy of code-based generation. Their visual quality is competitive with open-source diffusion (FID: 98, HVA: 4.0/5) but introduces a cost dimension that is untenable for educational deployment:

\begin{table}[t]
\caption{Cost analysis of closed-source diagram generation at educational scale. Assumes 12 diagrams per deck.}
\label{tab:cost}
\setlength{\tabcolsep}{4pt}
\begin{tabular}{@{}lrrr@{}}
\toprule
\textbf{Scenario} & \textbf{DALL\textperiodcentered E~3} & \textbf{GPT-4o} & \textbf{Gemini} \\
\midrule
Per image (\$) & 0.040 & 0.080 & 0.040 \\
Per deck (\$) & 0.48 & 0.96 & 0.48 \\
Teacher/yr\textsuperscript{a} (\$) & 19.20 & 38.40 & 19.20 \\
School/yr\textsuperscript{b} (\$) & 960 & 1{,}920 & 960 \\
\midrule
Eff.\ cost\textsuperscript{c} (\$) & \multicolumn{3}{c}{1.3--1.6$\times$ above} \\
\bottomrule
\end{tabular}
\vspace{2pt}

\raggedright\scriptsize
\textsuperscript{a}40 wks $\times$ 1 deck/wk.
\textsuperscript{b}50 teachers.
\textsuperscript{c}${\sim}$30\% of images require re-generation at observed LEM.
\end{table}

Beyond cost, closed-source models present four additional barriers for educational and research use: (1)~\emph{reproducibility}---proprietary model weights and undisclosed training data preclude replication; (2)~\emph{content filtering}---aggressive safety filters occasionally reject legitimate educational content (e.g., anatomical diagrams); (3)~\emph{dependency risk}---pricing changes or API deprecation can invalidate deployed educational workflows overnight; and (4)~\emph{compliance}---sending student-related data to third-party APIs raises concerns under regulations such as COPPA~\cite{ritvo2013privacy} and FERPA.

\subsection{Summary: The Case for a New Paradigm}

\begin{table}[t]
\caption{Accuracy and visual quality across paradigms on 400 K--12 diagram prompts. LEM\,=\,Label Exact-Match (\%); CER\,=\,Character\ Error Rate (\%); FID\,=\,Fr\'{e}chet Inception Distance (against textbook reference, $\downarrow$); HVA\,=\,Human Visual Appeal (1--5, $\uparrow$; Krippendorff's $\alpha = 0.78$, $n{=}6$ annotators).}
\label{tab:accuracy}
\setlength{\tabcolsep}{2.5pt}
\begin{tabular}{@{}llccccc@{}}
\toprule
\textbf{Paradigm} & \textbf{Model} & \textbf{LEM}$\uparrow$ & \textbf{CER}$\downarrow$ & \textbf{FID}$\downarrow$ & \textbf{HVA}$\uparrow$ & \textbf{\$/img} \\
\midrule
\multirow{3}{*}{\shortstack[l]{Open-src.\\diffusion}}
 & SDXL 1.0      & 11.3 & 71.4 & 112.6 & 3.8 & 0 \\
 & Flux.1-dev     & 18.7 & 59.2 & 95.3 & 4.1 & 0 \\
 & SD3 Med.       & 14.9 & 64.8 & 103.7 & 3.9 & 0 \\
\midrule
\multirow{2}{*}{\shortstack[l]{Code-based\\(LLM$\to$code)}}
 & GPT-4o         & 97.2 & 0.8 & 184.5 & 2.1 & 0\textsuperscript{a} \\
 & Claude 3.5     & 95.6 & 1.4 & 191.3 & 2.0 & 0\textsuperscript{a} \\
\midrule
\multirow{3}{*}{\shortstack[l]{Closed-src.\\APIs}}
 & DALL\textperiodcentered E~3       & 64.3 & 19.7 & 98.4 & 4.0 & 0.04 \\
 & GPT-4o img     & 73.8 & 14.2 & 91.2 & 4.2 & 0.08 \\
 & Gemini         & 59.1 & 23.6 & 105.8 & 3.8 & 0.04 \\
\midrule
Ours & CAGE           & 92.4 & 2.6 & 97.1 & 3.9 & 0 \\
\bottomrule
\end{tabular}
\vspace{2pt}

\raggedright\scriptsize
\textsuperscript{a}Code executed locally; LLM API cost excluded (shared across all LLM-based approaches).
\end{table}

Table~\ref{tab:accuracy} crystallizes the dilemma across both axes. A clear Pareto frontier emerges: code-based methods dominate on accuracy (highest LEM, lowest CER) but score worst on visual quality (highest FID, lowest HVA); open-source diffusion models show the inverse pattern, excelling on aesthetics while failing catastrophically on label fidelity. Closed-source APIs approach both frontiers but at a cost---both financial and methodological---that is incompatible with scalable educational deployment and open research. CAGE is the only approach that simultaneously achieves high label accuracy and strong visual quality at zero marginal per-image cost.

This gap is not merely inconvenient; it is \emph{structurally inevitable} under current single-stage approaches. Diffusion models generate images holistically in a compressed latent space where character-level precision is fundamentally at odds with the generative objective~\cite{ma2023glyphdraw}. Code-based rendering is precise because it operates in a symbolic space, but this same symbolic representation precludes the continuous, texture-rich outputs that diffusion excels at. The insight underlying our proposal is that these limitations are \emph{complementary}: what code does well (accuracy), diffusion does poorly, and vice versa.

\section{CAGE: Code-Anchored Generative Enhancement}
\label{sec:method}

We propose \emph{CAGE} (Code-Anchored Generative Enhancement), a two-stage paradigm that decouples the accuracy and aesthetics concerns in educational diagram generation. The key insight is simple: let code handle what code is good at (precision), and let diffusion handle what diffusion is good at (visual quality), rather than asking either to do both.

\subsection{Overview}

Given a natural language description $d$ of an educational diagram (e.g., ``a labeled diagram of the Krebs cycle showing pyruvate, acetyl-CoA, citrate, \ldots''), the pipeline proceeds in two stages:

\begin{enumerate}
    \item \textbf{Stage 1: Constrained Code Synthesis.} An LLM $\mathcal{L}$ receives $d$ and generates executable code $c = \mathcal{L}(d)$ in Python (Matplotlib, NetworkX), \LaTeX{}/TikZ, or SVG. Executing $c$ yields a programmatic rendering $I_{\text{prog}}$ with guaranteed label accuracy and structural correctness.

    \item \textbf{Stage 2: Structure-Preserving Diffusion Refinement.} A diffusion model $\mathcal{D}$, conditioned on structural features extracted from $I_{\text{prog}}$, generates a visually polished output $I_{\text{ref}} = \mathcal{D}(I_{\text{prog}}, s)$ where $s$ is a style prompt (e.g., ``professional educational illustration, textbook quality''). Structural conditioning ensures that the layout, connectivity, and label positions of $I_{\text{prog}}$ are preserved in $I_{\text{ref}}$.
\end{enumerate}

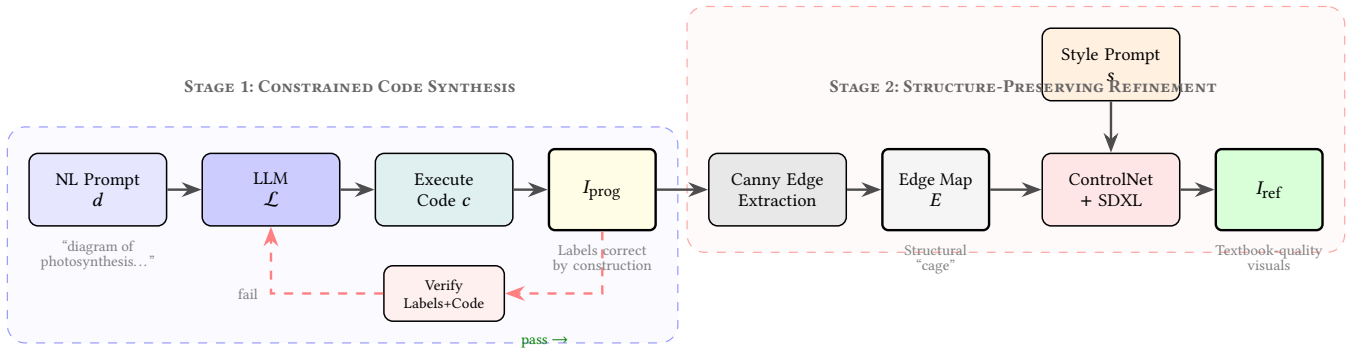
\begin{figure*}[t]
    \centering
    \resizebox{\textwidth}{!}{%
    \begin{tikzpicture}[
        >=Stealth,
        node distance=0.3cm and 0.35cm,
        block/.style={draw, rounded corners=3pt, minimum height=0.85cm, text width=1.4cm, align=center, font=\scriptsize, line width=0.5pt},
        img/.style={draw, thick, minimum height=0.95cm, minimum width=1.25cm, align=center, font=\scriptsize, rounded corners=2pt, line width=0.8pt},
        arr/.style={->, thick, color=black!70},
        stglabel/.style={font=\scriptsize\bfseries, text=black!60},
        annot/.style={font=\tiny, text=black!50, align=center},
    ]
    \node[block, fill=blue!10] (prompt) {NL Prompt\\$d$};
    \node[block, fill=blue!20, right=0.4cm of prompt] (llm) {LLM\\$\mathcal{L}$};
    \node[block, fill=teal!12, right=0.4cm of llm] (exec) {Execute\\Code $c$};
    \node[img, fill=yellow!12, right=0.4cm of exec] (iprog) {$I_{\text{prog}}$};

    \node[block, fill=gray!18, right=0.6cm of iprog] (canny) {Canny Edge\\Extraction};
    \node[img, fill=gray!8, right=0.4cm of canny] (edges) {Edge Map\\$E$};

    \node[block, fill=red!10, right=0.6cm of edges] (cn) {ControlNet\\$+$ SDXL};
    \node[img, fill=green!15, right=0.4cm of cn] (iref) {$I_{\text{ref}}$};

    \node[block, fill=orange!12, above=0.6cm of cn] (style) {Style Prompt\\$s$};

    \node[block, fill=red!6, below=0.45cm of exec, text width=1.2cm, minimum height=0.65cm] (verify) {\tiny Verify\\[-1pt]\tiny Labels+Code};

    \draw[arr] (prompt) -- (llm);
    \draw[arr] (llm) -- (exec);
    \draw[arr] (exec) -- (iprog);
    \draw[arr] (iprog) -- (canny);
    \draw[arr] (canny) -- (edges);
    \draw[arr] (edges) -- (cn);
    \draw[arr] (style) -- (cn);
    \draw[arr] (cn) -- (iref);

    \draw[arr, dashed, color=red!50] (iprog.south) |- (verify.east);
    \draw[arr, dashed, color=red!50] (verify.west) -| node[annot, left, xshift=-1pt] {fail} (llm.south);
    \node[annot, below right=0.06cm and 0.08cm of verify.south east, text=green!50!black] {pass $\rightarrow$};

    \node[annot, below=0.04cm of prompt] {``diagram of\\[-1pt]photosynthesis\ldots''};
    \node[annot, below=0.04cm of iprog, text width=1.3cm] {Labels correct\\[-1pt]by construction};
    \node[annot, below=0.04cm of edges, text width=1.1cm] {Structural\\[-1pt]``cage''};
    \node[annot, below=0.04cm of iref, text width=1.3cm] {Textbook-quality\\[-1pt]visuals};

    \node[stglabel] at ($(prompt.north)!0.5!(iprog.north)+(0,0.75)$) {\textsc{Stage 1: Constrained Code Synthesis}};
    \node[stglabel] at ($(canny.north)!0.5!(iref.north)+(0,0.75)$) {\textsc{Stage 2: Structure-Preserving Refinement}};

    \begin{scope}[on background layer]
        \node[draw=blue!40, dashed, rounded corners=6pt, inner sep=7pt, fit=(prompt)(llm)(exec)(iprog)(verify), fill=blue!2] {};
        \node[draw=red!35, dashed, rounded corners=6pt, inner sep=7pt, fit=(canny)(edges)(cn)(iref)(style), fill=red!2] {};
    \end{scope}
    \end{tikzpicture}%
    }
    \caption{The CAGE pipeline. Stage~1 synthesizes executable code via an LLM, executes it to produce a structurally correct programmatic rendering $I_{\text{prog}}$, and verifies label completeness in a REPL-style loop. Stage~2 extracts Canny edges from $I_{\text{prog}}$ as a spatial ``cage,'' then applies ControlNet-conditioned SDXL diffusion to produce $I_{\text{ref}}$---a visually polished output that preserves all labels and diagram topology.}
    \label{fig:pipeline}
\end{figure*}

\subsection{Stage 1: Constrained Code Synthesis}

The first stage leverages the established capability of LLMs to generate executable code~\cite{tian2024chartgpt, narechania2020nl4dv}. We frame the task as \emph{constrained} code synthesis: the generated code must satisfy structural invariants beyond mere executability.

\textbf{Why code guarantees accuracy.} When an LLM generates the Python statement \texttt{plt.text(x,\,y,\,"Mitochondria")}, the string ``Mitochondria'' is rendered by the font engine exactly as specified. There is no stochastic sampling, no latent-space compression, and no learned approximation of character shapes. The label is correct \emph{by construction}. Similarly, structural relationships---an arrow from ``Reactants'' to ``Products,'' a containment boundary around ``Nucleus''---are expressed as explicit geometric instructions rather than inferred from pixel statistics.

\textbf{Code verification.} The programmatic rendering step provides a natural verification checkpoint. Before proceeding to Stage~2, we can apply automated checks: (1)~all ground-truth labels appear in the code's text-rendering calls; (2)~the code executes without errors; (3)~structural constraints (e.g., graph connectivity) are satisfied. Failed checks trigger re-generation with feedback, following a REPL-style correction loop similar to that employed by PPTAgent~\cite{zheng2025pptagent}.

\textbf{Language choice.} Different diagram types benefit from different rendering languages. Flowcharts and process diagrams map naturally to Graphviz or Mermaid; mathematical illustrations benefit from TikZ's coordinate-based precision; biological diagrams with complex shapes are best served by Matplotlib or custom SVG generation. The choice of rendering language is itself a decision the LLM can make based on the prompt content.

\subsection{Stage 2: Structure-Preserving Diffusion Refinement}

The second stage transforms the visually flat programmatic rendering into a polished educational graphic. The critical constraint is \emph{structural preservation}: the diffusion model must enhance visual quality without altering the diagram's topology, label positions, or textual content.

\textbf{ControlNet and Canny edge conditioning.} We employ ControlNet~\cite{zhang2023adding} with Canny edge maps as the primary structural conditioning mechanism. The programmatic rendering $I_{\text{prog}}$ is converted to a Canny edge map $E = \text{Canny}(I_{\text{prog}})$, which captures the precise outlines of boxes, arrows, boundaries, and label positions. This edge map serves as a spatial ``cage'' that constrains the diffusion process: the model can alter colors, textures, shading, and fine details, but cannot move structural elements because the edge conditioning provides pixel-level positional constraints~\cite{zhang2023adding}.

This choice is deliberate. Among available conditioning mechanisms listed below, Canny edges provide the strongest structural guarantee for diagram refinement:
\begin{itemize}
    \item \emph{SDEdit}~\cite{meng2022sdedit} applies global noise and denoise, which preserves coarse structure but destroys text at high noise levels.
    \item \emph{InstructPix2Pix}~\cite{brooks2023instructpix2pix} supports targeted edits but offers ``soft'' structural preservation that is insufficient for diagram topology.
    \item \emph{ControlNet with Canny edges}~\cite{zhang2023adding} provides ``hard'' spatial control, enforcing pixel-level boundary adherence---precisely what diagram refinement requires.
\end{itemize}

\textbf{Label preservation strategy.} Even with ControlNet conditioning, text regions require special handling. We employ a masking-and-recomposition approach: (1)~detect text regions in $I_{\text{prog}}$ via OCR bounding boxes; (2)~mask these regions during the diffusion refinement process; (3)~after stylization, re-render the labels using a font style matched to the diffusion output's aesthetic. This ensures that labels remain pixel-perfect regardless of the diffusion model's text-rendering capability.

\textbf{Style control.} The style prompt `$s$' guides the aesthetic transformation. For K--12 contexts, we use prompts emphasizing clarity and engagement (e.g., ``clean educational illustration, professional textbook diagram, clear colors, white background''). The style can be adapted to grade-level appropriateness: younger students benefit from bolder, simpler renderings, while advanced courses warrant more detailed, technical aesthetics. We show the complete pipleline in Figure \ref{fig:pipeline}.

\section{EduDiagram-2K Dataset}
\label{sec:dataset}

A critical enabler of the CAGE paradigm is the availability of paired training data mapping programmatic diagram renderings to their stylized educational counterparts. Our review of the educational data landscape reveals a conspicuous absence of such a resource. Existing visual datasets---FigureQA~\cite{kahou2017figureqa}, PlotQA~\cite{methani2020plotqa}, ChartQA~\cite{masry2022chartqa}---focus on chart \emph{understanding} (extracting data from existing visualizations) rather than diagram \emph{generation} or \emph{stylization}. SciGraphQA~\cite{li2023scigraphqa} addresses multi-turn reasoning over scientific graphs but does not provide paired raw-to-pedagogical mappings. Educational QA datasets such as TQA~\cite{kembhavi2017you} and ScienceQA~\cite{lu2022learn} include diagrams as inputs for comprehension tasks but do not address diagram \emph{creation}. No existing dataset supports the task of transforming a programmatic diagram into a visually polished educational graphic while preserving structural fidelity.

\subsection{Dataset Construction}

\begin{figure*}[!htp]
    \centering
    \setlength{\tabcolsep}{4pt}
    \begin{tabular}{cc}
    {\small\bfseries $I_{\text{prog}}$ (code-generated)} & {\small\bfseries $I_{\text{style}}$ (CAGE-refined)} \\[4pt]
    \includegraphics[width=0.48\textwidth]{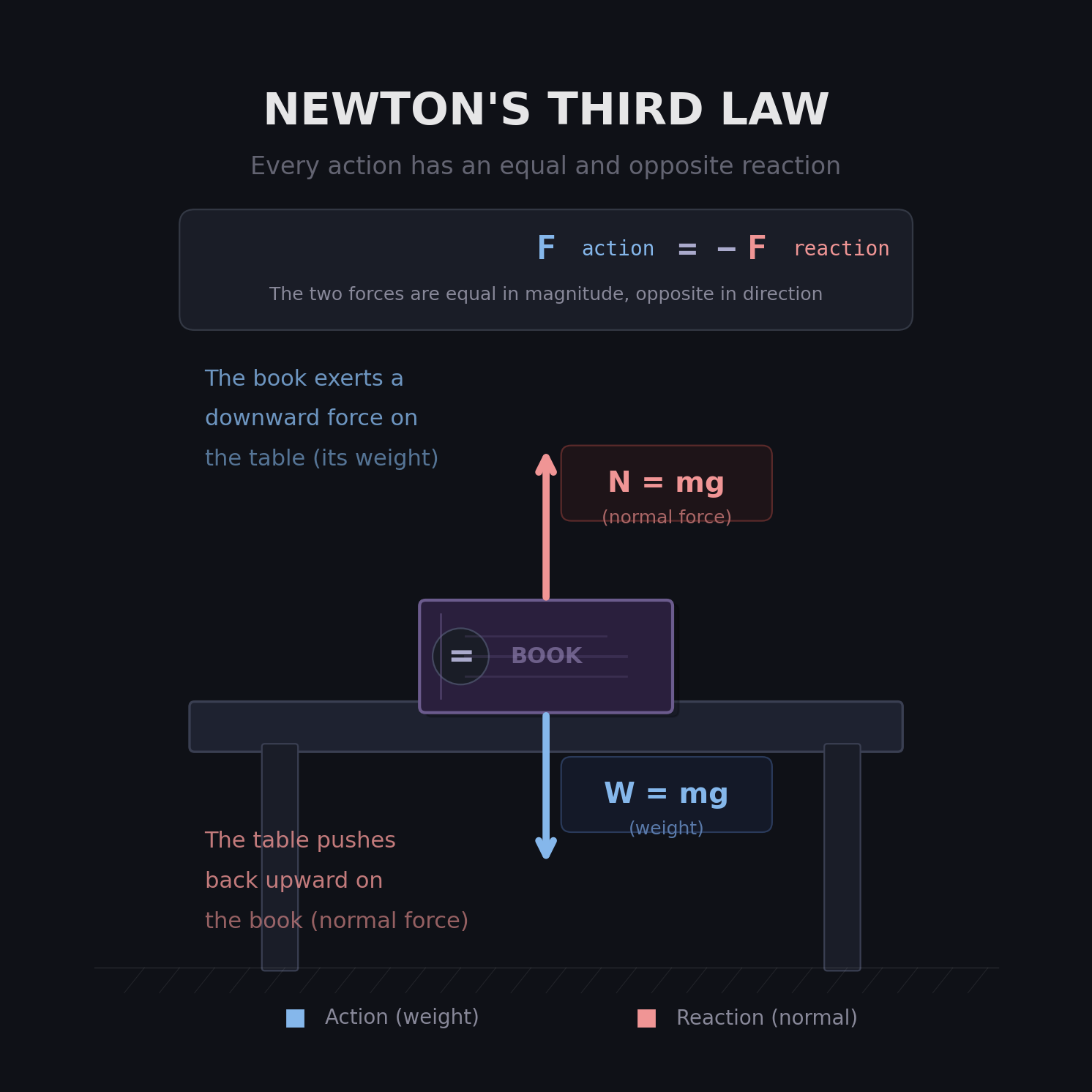} &
    \includegraphics[width=0.48\textwidth]{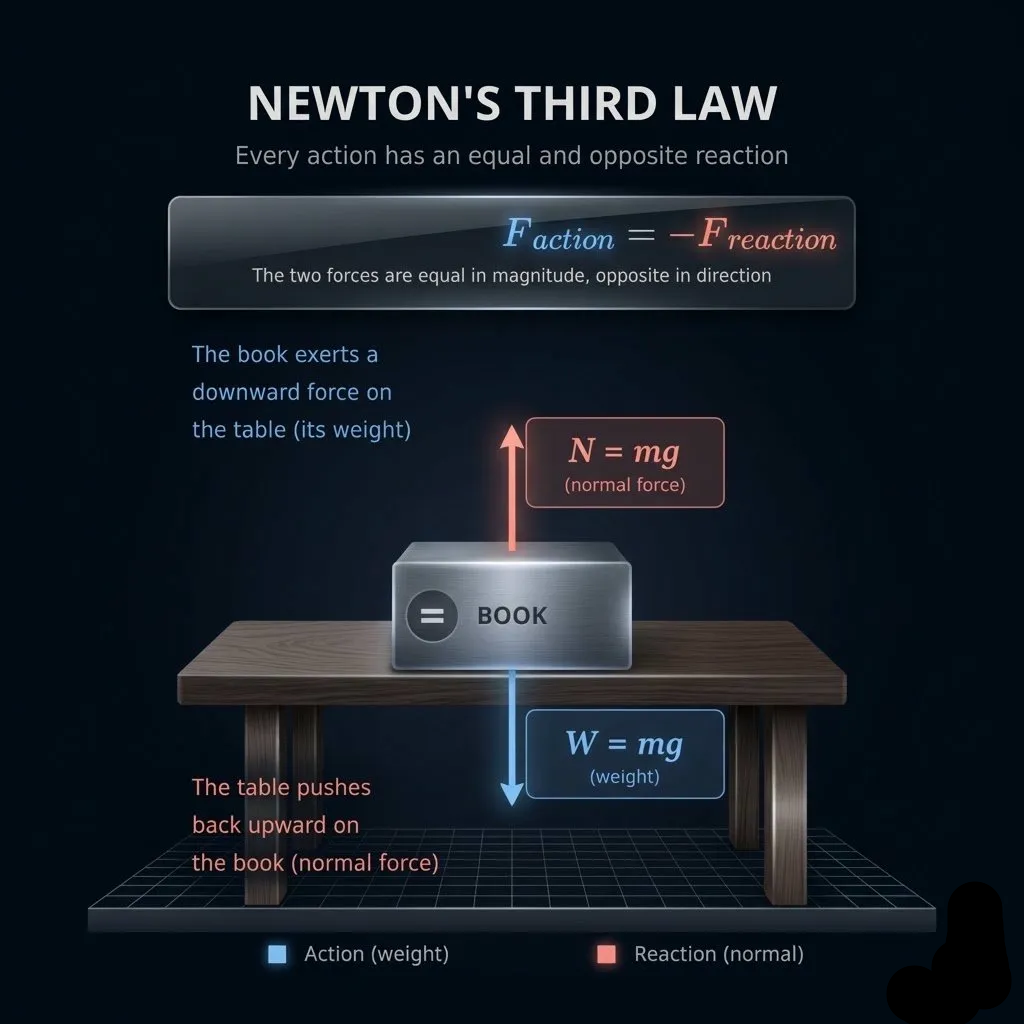} \\[-1pt]
    \multicolumn{2}{c}{\small (a) Newton's third law of motion} \\[6pt]
    \includegraphics[width=0.48\textwidth]{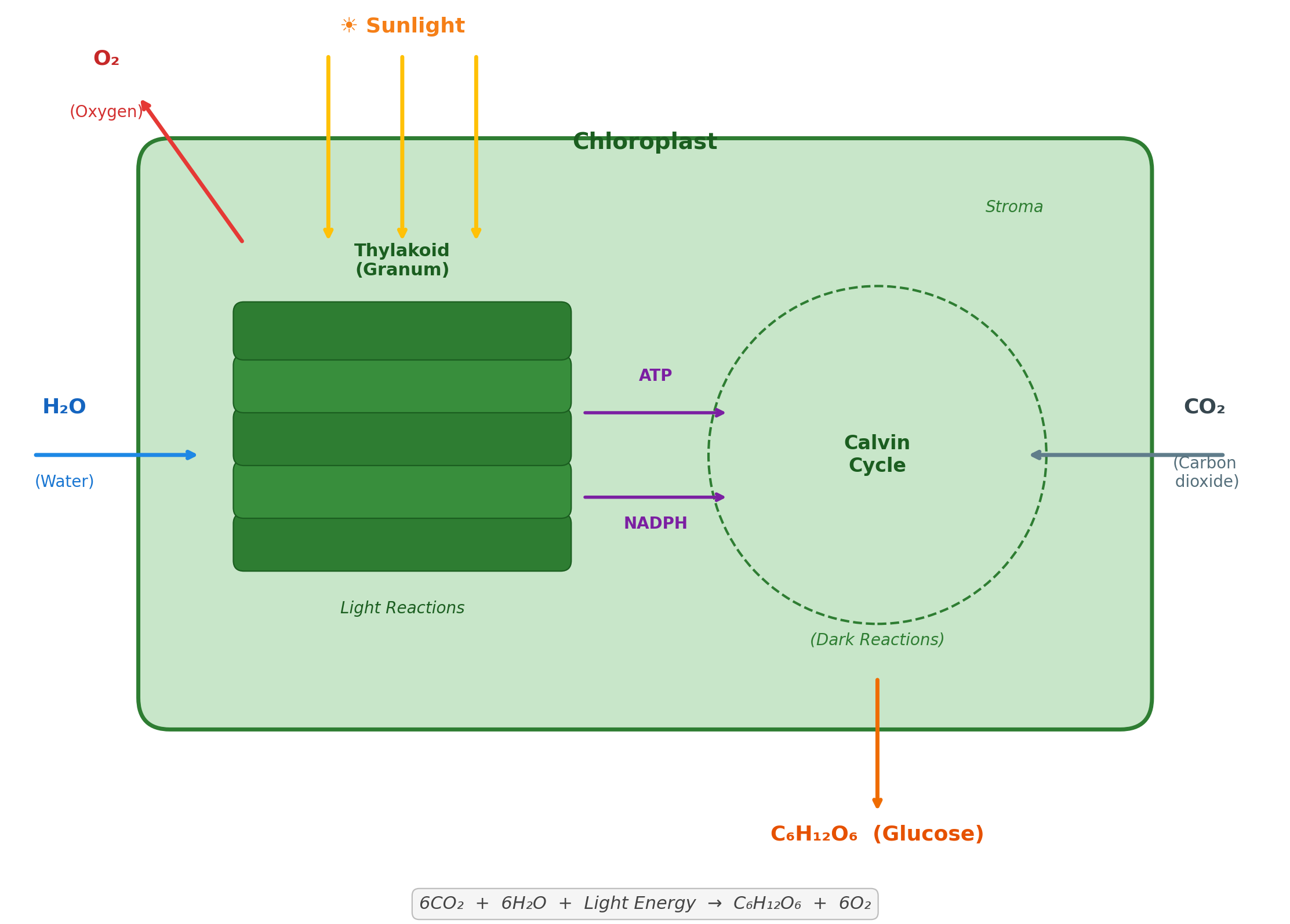} &
    \includegraphics[width=0.48\textwidth]{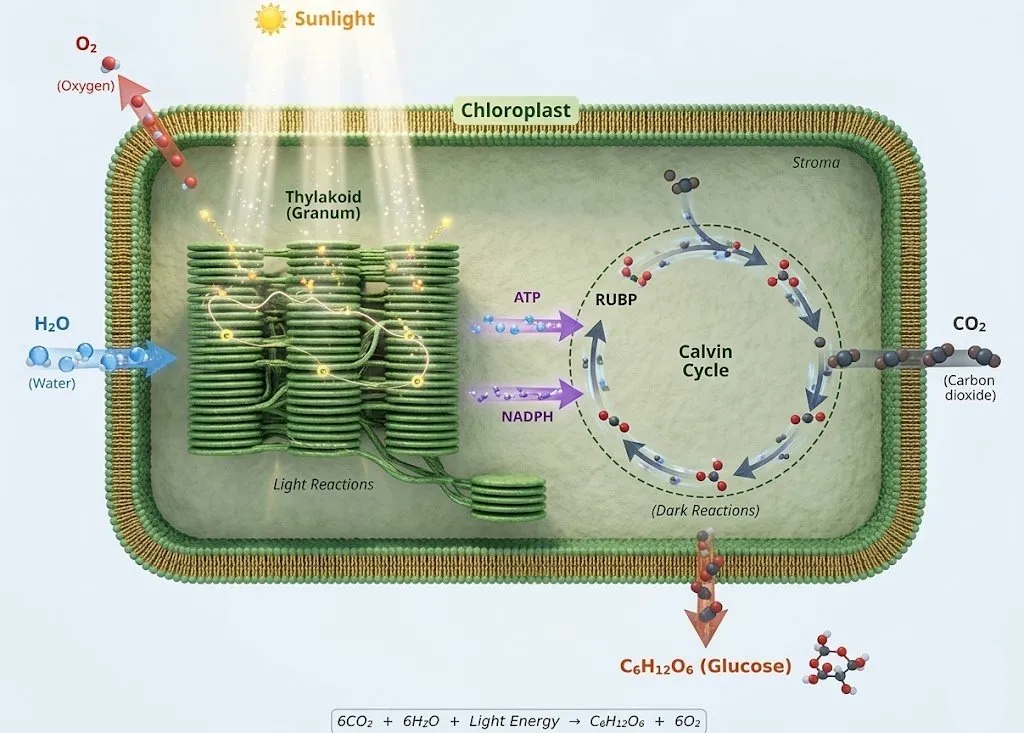} \\[-1pt]
    \multicolumn{2}{c}{\small (b) Photosynthesis} \\
    \end{tabular}
    \caption{Sample pairs from EduDiagram-2K. Each row shows a code-generated diagram ($I_{\text{prog}}$, left) and its CAGE-refined stylized version ($I_{\text{style}}$, right). All labels and structural topology are preserved while visual quality is substantially enhanced.}
    \label{fig:dataset}
\end{figure*}

EduDiagram-2K consists of approximately 2,000 paired examples $(I_{\text{prog}}, I_{\text{style}})$, where $I_{\text{prog}}$ is a programmatically generated diagram and $I_{\text{style}}$ is its stylized educational counterpart as shown in Figure \ref{fig:dataset}. Construction proceeds in three phases.

\textbf{Phase 1: Prompt curation.} We curate diagram prompts from K--12 STEM curriculum aligned with widely adopted standards (NGSS for science, Common Core for mathematics). Each prompt specifies a topic, a set of required labels, and a target grade band. Prompts are stratified across subjects and grade levels to ensure broad coverage.

\textbf{Phase 2: Programmatic generation.} For each prompt, we use LLM-driven code synthesis (GPT-4o and Claude 3.5 Sonnet) to generate executable code that renders the requested diagram. We employ multiple rendering backends---Matplotlib and NetworkX for scientific diagrams, \LaTeX{}/TikZ for mathematical illustrations, and custom SVG for spatial layouts---selecting the backend that best matches the diagram type. Each generated $I_{\text{prog}}$ is verified for: (a)~executability (code runs without errors), (b)~label completeness (all required labels appear in rendering calls), and (c)~structural plausibility (manual spot-check of spatial relationships). Failed generations are re-prompted with error feedback.

\textbf{Phase 3: Stylized counterpart construction.} The stylized counterparts $I_{\text{style}}$ are produced through a semi-automated pipeline with human oversight. We apply the CAGE refinement process (ControlNet with Canny edge conditioning) to each $I_{\text{prog}}$, generating candidate stylizations at multiple style strengths. A team of annotators with educational publishing experience then selects or rejects each candidate based on three criteria: (1)~all labels from $I_{\text{prog}}$ are preserved exactly (verified via OCR comparison); (2)~the structural topology is maintained (node connectivity, containment, spatial ordering); (3)~the visual quality meets educational publishing standards (clear colors, professional appearance, age-appropriate complexity). Rejected candidates are re-generated with adjusted style prompts or manually corrected. Approximately 68\% of candidates pass verification on the first attempt.

\subsection{Dataset Statistics}

Table~\ref{tab:dataset} summarizes the key properties of EduDiagram-2K. The dataset spans four STEM subjects across three grade bands, with an average of 8.4 labels per diagram.

\begin{table}[t]
\caption{EduDiagram-2K dataset statistics.}
\label{tab:dataset}
\begin{tabular}{@{}lp{4.2cm}@{}}
\toprule
\textbf{Property} & \textbf{Value} \\
\midrule
Total pairs & $\sim$2{,}000 \\
Subjects & Biology, Chemistry, Physics, Mathematics \\
Grade bands & K--5, 6--8, 9--12 \\
Diagram types & Flowcharts, labeled illustrations, concept maps, structural diagrams \\
Avg.\ labels/diagram & 8.4 \\
Rendering languages & Python, \LaTeX{}/TikZ, SVG \\
\bottomrule
\end{tabular}
\end{table}

\subsection{Novelty and Significance}

EduDiagram-2K is, to our knowledge, the first dataset providing paired programmatic-to-stylized educational diagrams. Its significance extends beyond the CAGE pipeline:

\begin{itemize}
    \item \textbf{Benchmarking structural fidelity.} The paired format enables quantitative evaluation of structure-preserving image translation methods using metrics such as IoU of bounding boxes and label exact-match rates.
    \item \textbf{Training diffusion models.} The paired data can be used to fine-tune ControlNet or InstructPix2Pix models specifically for the educational diagram domain, reducing the stylistic gap between programmatic and professional renderings.
    \item \textbf{Educational content generation.} The dataset establishes a concrete standard for what ``textbook-quality'' educational diagrams look like across subjects and grade levels.
\end{itemize}

\section{Research Agenda and Open Problems}
\label{sec:agenda}

The CAGE paradigm, while demonstrated here in proof-of-concept form, opens several foundational research problems that we believe are ripe for investigation by the multimedia community.

\textbf{Grade-adaptive stylization.} Educational diagrams for a kindergartener learning about plant growth must look fundamentally different from those for a high-school student studying cellular respiration. The visual complexity, color palette, level of anatomical detail, and label density should all be adapted to the target grade level. Current diffusion models have no notion of ``grade-appropriate'' visual complexity. Developing conditioning mechanisms or style spaces that encode pedagogical age-appropriateness is an open multimodal learning problem.

\textbf{Curriculum-grounded code synthesis.} Stage~1 of our pipeline currently relies on the LLM's parametric knowledge to determine \emph{what} to depict. For K--12 use, the content must be grounded in specific curriculum standards (e.g., NGSS, Common Core) and vetted textbook sources. Integrating retrieval-augmented generation with constrained code synthesis---ensuring that the LLM generates code for a diagram that is both factually correct and curriculum-aligned---is a challenging multimodal retrieval problem.

\textbf{Evaluation frameworks for educational diagrams.} Standard image quality metrics (FID, KID) are designed for photorealism, not pedagogical effectiveness. Existing evaluation frameworks such as PPTEval~\cite{zheng2025pptagent} assess presentation quality but do not address diagram-specific criteria, while EduVisBench~\cite{ji2025eduvisbench} provides rubrics for visualization literacy but not for generative diagram evaluation. Evaluating educational diagrams requires metrics that jointly assess textual accuracy, structural fidelity, visual engagement, and learning efficacy. The development of such metrics---potentially grounded in learning science frameworks such as Mayer's Cognitive Theory of Multimedia Learning~\cite{mayer2013multimedia, mayer2021evidence}---would benefit the broader educational multimedia community.

\textbf{Beyond static diagrams.} The CAGE paradigm naturally extends to animated and interactive educational visuals. A code-generated animation (e.g., a step-by-step walkthrough of mitosis) could be refined by video diffusion models to produce engaging educational animations. Similarly, interactive diagrams---where students can manipulate variables and observe changes---could benefit from diffusion-enhanced rendering of intermediate states.

\textbf{Multimodal educational content generation.} While this paper focuses specifically on diagram generation, the broader vision involves integrating accurate diagrams into full multimodal educational presentations---slides with synchronized narration, interactive assessments, and adaptive difficulty. The CAGE paradigm provides the visual component; complementary work on text-to-speech for educational narration~\cite{ren2021fastspeech} and retrieval-augmented slide generation could compose a complete, grounded educational content pipeline.

\textbf{Accessibility and multilingual adaptation.} Educational diagrams must be accessible to students with visual impairments (requiring alt-text generation grounded in the code's structural representation) and adaptable to multiple languages (requiring label translation that preserves spatial layout). The code-based Stage~1 representation makes both of these adaptations more tractable than they would be with purely image-based generation.

\section{Conclusion}

We have presented empirical evidence that the generation of educational diagrams with accurate text labels faces a fundamental accuracy--aesthetics dilemma: diffusion models produce visually engaging but textually unreliable images, code-based generation ensures accuracy but lacks visual quality, and closed-source APIs offer a partial but costly and unreliable compromise. To resolve this dilemma, we propose CAGE---a two-stage paradigm that delegates accuracy to code synthesis and aesthetics to structure-preserving diffusion refinement. We introduce EduDiagram-2K, the first paired dataset of programmatic and stylized educational diagrams, as a concrete resource enabling this paradigm. Our evaluation across 400 K--12 diagram prompts---measuring label accuracy via LEM and CER, and visual quality via FID and human visual appeal ratings---confirms that CAGE is the first approach to simultaneously achieve high fidelity on both axes at zero marginal per-image cost. We invite the multimedia community to build on CAGE and the EduDiagram-2K dataset to advance the state of accurate, engaging, and accessible educational visual content.

\bibliographystyle{ACM-Reference-Format}
\bibliography{references}

@incollection{mayer2013multimedia,
  title={Multimedia instruction},
  author={Mayer, Richard E},
  booktitle={Handbook of research on educational communications and technology},
  pages={385--399},
  year={2013},
  publisher={Springer}
}

@article{mcallister2026understanding,
  title={Understanding K-12 Public High School Teachers’ Perceptions of Artificial Intelligence in Education: A Phenomenological Study},
  author={McAllister, Jackie Samantha},
  year={2026}
}

@inproceedings{zheng2025pptagent,
  title={Pptagent: Generating and evaluating presentations beyond text-to-slides},
  author={Zheng, Hao and Guan, Xinyan and Kong, Hao and Zhang, Wenkai and Zheng, Jia and Zhou, Weixiang and Lin, Hongyu and Lu, Yaojie and Han, Xianpei and Sun, Le},
  booktitle={Proceedings of the 2025 Conference on Empirical Methods in Natural Language Processing},
  pages={14413--14429},
  year={2025}
}

@article{liang2025slidegen,
  title={Slidegen: Collaborative multimodal agents for scientific slide generation},
  author={Liang, Xin and Zhang, Xiang and Xu, Yiwei and Sun, Siqi and You, Chenyu},
  journal={arXiv preprint arXiv:2512.04529},
  year={2025}
}

@inproceedings{bandyopadhyay2024enhancing,
  title={Enhancing presentation slide generation by llms with a multi-staged end-to-end approach},
  author={Bandyopadhyay, Sambaran and Maheshwari, Himanshu and Natarajan, Anandhavelu and Saxena, Apoorv},
  booktitle={Proceedings of the 17th International Natural Language Generation Conference},
  pages={222--229},
  year={2024}
}

@article{podell2023sdxl,
  title={Sdxl: Improving latent diffusion models for high-resolution image synthesis},
  author={Podell, Dustin and English, Zion and Lacey, Kyle and Blattmann, Andreas and Dockhorn, Tim and M{\"u}ller, Jonas and Penna, Joe and Rombach, Robin},
  journal={arXiv preprint arXiv:2307.01952},
  year={2023}
}

@article{greenberg2025demystifying,
  title={Demystifying flux architecture},
  author={Greenberg, Or},
  journal={arXiv preprint arXiv:2507.09595},
  year={2025}
}

@inproceedings{esser2024scaling,
  title={Scaling rectified flow transformers for high-resolution image synthesis},
  author={Esser, Patrick and Kulal, Sumith and Blattmann, Andreas and Entezari, Rahim and M{\"u}ller, Jonas and Saini, Harry and Levi, Yam and Lorenz, Dominik and Sauer, Axel and Boesel, Frederic and others},
  booktitle={Forty-first international conference on machine learning},
  year={2024}
}

@article{betker2023improving,
  title={Improving image generation with better captions},
  author={Betker, James and Goh, Gabriel and Jing, Li and Brooks, Tim and Wang, Jianfeng and Li, Linjie and Ouyang, Long and Zhuang, Juntang and Lee, Joyce and Guo, Yufei and others},
  journal={Computer Science. https://cdn. openai. com/papers/dall-e-3. pdf},
  volume={2},
  number={3},
  pages={8},
  year={2023}
}

@article{ma2023glyphdraw,
  title={Glyphdraw: Seamlessly rendering text with intricate spatial structures in text-to-image generation},
  author={Ma, Jian and Zhao, Mingjun and Chen, Chen and Wang, Ruichen and Niu, Di and Lu, Haonan and Lin, Xiaodong},
  journal={arXiv preprint arXiv:2303.17870},
  year={2023}
}

@inproceedings{zhang2024brush,
  title={Brush your text: Synthesize any scene text on images via diffusion model},
  author={Zhang, Lingjun and Chen, Xinyuan and Wang, Yaohui and Lu, Yue and Qiao, Yu},
  booktitle={Proceedings of the AAAI Conference on Artificial Intelligence},
  volume={38},
  number={7},
  pages={7215--7223},
  year={2024}
}

@article{tian2024chartgpt,
  title={Chartgpt: Leveraging llms to generate charts from abstract natural language},
  author={Tian, Yuan and Cui, Weiwei and Deng, Dazhen and Yi, Xinjing and Yang, Yurun and Zhang, Haidong and Wu, Yingcai},
  journal={IEEE Transactions on Visualization and Computer Graphics},
  volume={31},
  number={3},
  pages={1731--1745},
  year={2024},
  publisher={IEEE}
}

@article{narechania2020nl4dv,
  title={NL4DV: A toolkit for generating analytic specifications for data visualization from natural language queries},
  author={Narechania, Arpit and Srinivasan, Arjun and Stasko, John},
  journal={IEEE Transactions on Visualization and Computer Graphics},
  volume={27},
  number={2},
  pages={369--379},
  year={2020},
  publisher={IEEE}
}

@inproceedings{zhang2023adding,
  title={Adding conditional control to text-to-image diffusion models},
  author={Zhang, Lvmin and Rao, Anyi and Agrawala, Maneesh},
  booktitle={Proceedings of the IEEE/CVF international conference on computer vision},
  pages={3836--3847},
  year={2023}
}

@inproceedings{
meng2022sdedit,
title={{SDE}dit: Guided Image Synthesis and Editing with Stochastic Differential Equations},
author={Chenlin Meng and Yutong He and Yang Song and Jiaming Song and Jiajun Wu and Jun-Yan Zhu and Stefano Ermon},
booktitle={International Conference on Learning Representations},
year={2022},
url={https://openreview.net/forum?id=aBsCjcPu_tE}
}

@inproceedings{brooks2023instructpix2pix,
  title={Instructpix2pix: Learning to follow image editing instructions},
  author={Brooks, Tim and Holynski, Aleksander and Efros, Alexei A},
  booktitle={Proceedings of the IEEE/CVF conference on computer vision and pattern recognition},
  pages={18392--18402},
  year={2023}
}

@article{kahou2017figureqa,
  title={Figureqa: An annotated figure dataset for visual reasoning},
  author={Kahou, Samira Ebrahimi and Michalski, Vincent and Atkinson, Adam and K{\'a}d{\'a}r, {\'A}kos and Trischler, Adam and Bengio, Yoshua},
  journal={arXiv preprint arXiv:1710.07300},
  year={2017}
}

@inproceedings{methani2020plotqa,
  title={Plotqa: Reasoning over scientific plots},
  author={Methani, Nitesh and Ganguly, Pritha and Khapra, Mitesh M and Kumar, Pratyush},
  booktitle={Proceedings of the ieee/cvf winter conference on applications of computer vision},
  pages={1527--1536},
  year={2020}
}

@inproceedings{masry2022chartqa,
  title={Chartqa: A benchmark for question answering about charts with visual and logical reasoning},
  author={Masry, Ahmed and Do, Xuan Long and Tan, Jia Qing and Joty, Shafiq and Hoque, Enamul},
  booktitle={Findings of the association for computational linguistics: ACL 2022},
  pages={2263--2279},
  year={2022}
}

@article{li2023scigraphqa,
  title={Scigraphqa: A large-scale synthetic multi-turn question-answering dataset for scientific graphs},
  author={Li, Shengzhi and Tajbakhsh, Nima},
  journal={arXiv preprint arXiv:2308.03349},
  year={2023}
}

@inproceedings{
ren2021fastspeech,
title={FastSpeech 2: Fast and High-Quality End-to-End Text to Speech},
author={Yi Ren and Chenxu Hu and Xu Tan and Tao Qin and Sheng Zhao and Zhou Zhao and Tie-Yan Liu},
booktitle={International Conference on Learning Representations},
year={2021},
url={https://openreview.net/forum?id=piLPYqxtWuA}
}

@article{mayer2021evidence,
  title={Evidence-based principles for how to design effective instructional videos},
  author={Mayer, Richard E},
  journal={Journal of Applied Research in Memory and Cognition},
  volume={10},
  number={2},
  pages={229--240},
  year={2021},
  publisher={Elsevier}
}

@article{ji2025eduvisbench,
  title={From eduvisbench to eduvisagent: A benchmark and multi-agent framework for reasoning-driven pedagogical visualization},
  author={Ji, Haonian and Qiu, Shi and Xin, Siyang and Han, Siwei and Chen, Zhaorun and Zhang, Dake and Wang, Hongyi and Yao, Huaxiu},
  journal={arXiv preprint arXiv:2505.16832},
  year={2025}
}

@article{luo2021natural,
  title={Natural language to visualization by neural machine translation},
  author={Luo, Yuyu and Tang, Nan and Li, Guoliang and Tang, Jiawei and Chai, Chengliang and Qin, Xuedi},
  journal={IEEE Transactions on Visualization and Computer Graphics},
  volume={28},
  number={1},
  pages={217--226},
  year={2021},
  publisher={IEEE}
}

@inproceedings{kembhavi2017you,
  title={Are you smarter than a sixth grader? textbook question answering for multimodal machine comprehension},
  author={Kembhavi, Aniruddha and Seo, Minjoon and Schwenk, Dustin and Choi, Jonghyun and Farhadi, Ali and Hajishirzi, Hannaneh},
  booktitle={Proceedings of the IEEE Conference on Computer Vision and Pattern recognition},
  pages={4999--5007},
  year={2017}
}

@article{lu2022learn,
  title={Learn to explain: Multimodal reasoning via thought chains for science question answering},
  author={Lu, Pan and Mishra, Swaroop and Xia, Tanglin and Qiu, Liang and Chang, Kai-Wei and Zhu, Song-Chun and Tafjord, Oyvind and Clark, Peter and Kalyan, Ashwin},
  journal={Advances in neural information processing systems},
  volume={35},
  pages={2507--2521},
  year={2022}
}

@inproceedings{sun2021d2s,
  title={D2S: Document-to-slide generation via query-based text summarization},
  author={Sun, Edward and Hou, Yufang and Wang, Dakuo and Zhang, Yunfeng and Wang, Nancy XR},
  booktitle={Proceedings of the 2021 Conference of the North American Chapter of the Association for Computational Linguistics: Human Language Technologies},
  pages={1405--1418},
  year={2021}
}

@article{chen2023textdiffuser,
  title={Textdiffuser: Diffusion models as text painters},
  author={Chen, Jingye and Huang, Yupan and Lv, Tengchao and Cui, Lei and Chen, Qifeng and Wei, Furu},
  journal={Advances in Neural Information Processing Systems},
  volume={36},
  pages={9353--9387},
  year={2023}
}

@inproceedings{rombach2022high,
  title={High-resolution image synthesis with latent diffusion models},
  author={Rombach, Robin and Blattmann, Andreas and Lorenz, Dominik and Esser, Patrick and Ommer, Bj{\"o}rn},
  booktitle={Proceedings of the IEEE/CVF conference on computer vision and pattern recognition},
  pages={10684--10695},
  year={2022}
}

@article{ritvo2013privacy,
  title={Privacy and Children's Data-An Overview of the Children's Online Privacy Protection Act and the Family Educational Rights and Privacy Act},
  author={Ritvo, Dalia and Bavitz, Christopher and Gupta, Ritu and Oberman, Irina},
  journal={Berkman Center Research Publication},
  number={23},
  year={2013}
}

@article{heusel2017gans,
  title={Gans trained by a two time-scale update rule converge to a local nash equilibrium},
  author={Heusel, Martin and Ramsauer, Hubert and Unterthiner, Thomas and Nessler, Bernhard and Hochreiter, Sepp},
  journal={Advances in neural information processing systems},
  volume={30},
  year={2017}
}

\end{document}